\documentclass{article}
\usepackage{spconf,amsmath,graphicx}
\usepackage{url}
\usepackage{booktabs}
\usepackage[numbers]{natbib}
\usepackage{amsmath, amsfonts, amsthm, stackrel, amssymb}

\begin{document}
\bibliographystyle{ieeetr}
\title{CAUSAL-STORY: LOCAL CAUSAL ATTENTION Utilizing Parameter-Efficient Tuning FOR VISUAL STORY SYNTHESIS
\vspace{-0.3cm}}
%
\name{
Tianyi Song\textsuperscript{1}, Jiuxin Cao*\textsuperscript{1,4}, Kun Wang\textsuperscript{1}, Bo Liu\textsuperscript{2}, Xiaofeng Zhang\textsuperscript{3}
\vspace{-0.2cm}}

\address{
\vspace{-0.1cm}
\textsuperscript{1}School of Cyber Science and Engineering, Southeast University, Nanjing, China \\
\vspace{-0.1cm}
\textsuperscript{2}School of Computer Science and Engineering, Southeast University, Nanjing, China \\
\vspace{-0.1cm}
\textsuperscript{3}Shanghai Jiao Tong University, Shanghai, China \\
\vspace{-0.2cm}
\textsuperscript{4}School of Computer Science and Engineering, Sanjiang University, Nanjing, China}
%
%

\maketitle
\begin{abstract}
The excellent text-to-image synthesis capability of diffusion models has driven progress in synthesizing coherent visual stories. The current state-of-the-art method combines the features of historical captions, historical frames, and the current captions as conditions for generating the current frame. However, this method treats each historical frame and caption as the same contribution. It connects them in order with equal weights, ignoring that not all historical conditions are associated with the generation of the current frame. To address this issue, we propose Causal-Story. This model incorporates a local causal attention mechanism that considers the causal relationship between previous captions, frames, and current captions. By assigning weights based on this relationship, Causal-Story generates the current frame, thereby improving the global consistency of story generation. We evaluated our model on the PororoSV and FlintstonesSV datasets and obtained state-of-the-art FID scores, and the generated frames also demonstrate better storytelling in visuals. 
\end{abstract}
%
\renewcommand{\thefootnote}{}
\footnotetext{*Corresponding author: (email) jx.cao@seu.edu.cn}
\section{Introduction}
\label{sec:intro}
Generating coherent visual narratives from natural language descriptions is a challenging task. It has far-reaching applications in fields such as story visualization, action prediction, and anime storyboard creation.

\begin{figure}[t]
\centering
\includegraphics[width=3.5in]{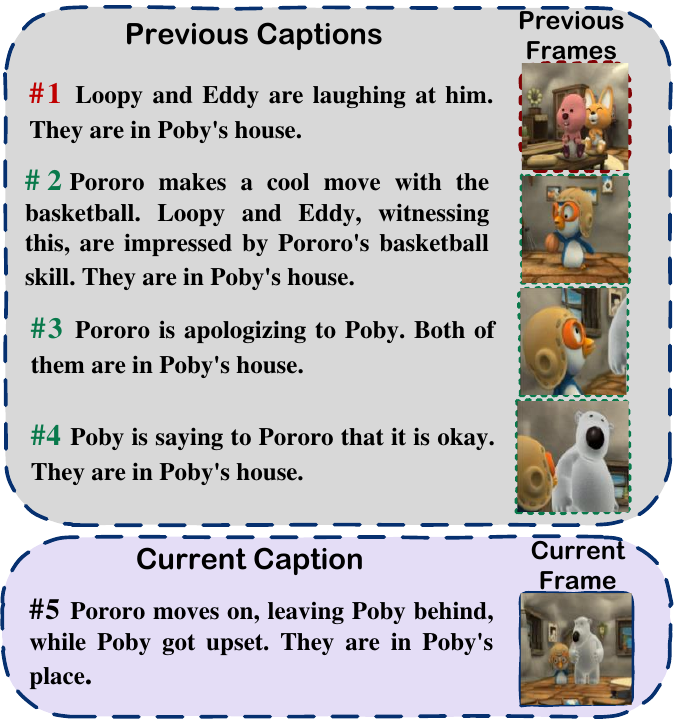}
\caption{An example of a story in PororoSV with five frames and captions. The green number indicates a dependency relationship between the previous frame and the current frame to be generated, while the red number indicates that it is not related to the generation of the current frame.}
\vspace{-0.5cm}
\label{introduction}
\end{figure}
Story visualization\cite{storygan} and story continuing\cite{storydalle} present a formidable challenge, necessitating the integration of contextual textual characteristics and historical frame details to yield convincing and coherent storylines with apt scene backgrounds and visual elements.
In coherent story synthesis, many parts are not covered by the caption of the current frame, such as objects, characters, actions, or backgrounds. This information may be contained in the description of several previous frames or included in the image features of the previous frames. 
For example, ``Loppy notice something. The woods are covered with snow.'' is the caption of the first frame. Moreover, ``Loppy explains what happened to Pororo's Flower'' is the caption of the second frame. Even though the environment is not described in the second caption, we know that the background of the second frame should include the woods covered with snow from the caption of the first frame.

Previous work \cite{vpcsv,storygan,vlcstorygan,wordlevelcsv,duco,vqvae,pororogan} mainly relied on Generative Adversarial Networks(GANs)\cite{gan}, autoregressive models\cite{aet2i}, and used contextual text encoders to improve consistency. However, these methods still need to improve in generating image quality and consistency. Maharana et al.\cite{storydalle} propose a new task setup for story continuation, using the first image as a condition. They fine-tune the large model DALL-E \cite{dalle} for the story visualization, which they call StoryDALL-E.

AR-LDM\cite{ARLDM} is a visual story generation model that builds upon the foundation of \cite{pre_arldm} by incorporating Stable Diffusion\cite{latentdiffusion}, which has enabled it to achieve the state-of-the-art FID score on benchmark datasets. Within the latent space, AR-LDM encodes the previous text-image context as a series of additional conditions\cite{clip}, in accordance with the chain rule. The UNet\cite{unet} decoder processes these additional conditions to produce the corresponding image.
One limitation of this approach is that it flattens all previous text-image pairs of the same story as conditioning memories, neglecting the fact that not all characters and scenes in the narrative are linearly connected.

Fig.1 illustrates that the generation of the fifth frame is predominantly influenced by the captions of the third and fourth frames, with no discernible correlation to the caption of the first frame. In contrast, the features extracted from the first frame may potentially impede the accurate generation of the fifth frame. We can measure their connection by the causal relationship between the corresponding textual captions of each frame.

We improved the model's attention mechanism, training, and sampling speed based on AR-LDM\cite{ARLDM}. Specifically, we make the following contributions:
\begin{enumerate}
\item We designed a local causal attention mask combined with latent diffusion to improve the model's judgment of contextual causal relationships.
\item We propose a lightweight adapter for efficient parameter tuning, which effectively reduces the training burden while ensuring training effectiveness. 
\item Quantitatively, we have achieved very competitive results on the PororoSV and FlintstonesSV test sets. Moreover, the training and inference speed has been improved under the same parameter. 
\end{enumerate}


\section{METHOD}
\label{sec:majhead}
In this section, we first formulate the probabilistic model of latent forward and reverse diffusion processes for consecutive story generation from text descriptions in2.1.
We then elaborate on the principles and mathematical expressions of causal attention mechanisms in 2.2. Finally, we introduce an adapter for efficient parameter tuning and the process of model training and inference in 2.3 and 2.4. Fig.2 illustrates the entire architecture.
\begin{figure*}[h]
  \centering
  \includegraphics[width=1.8\columnwidth]{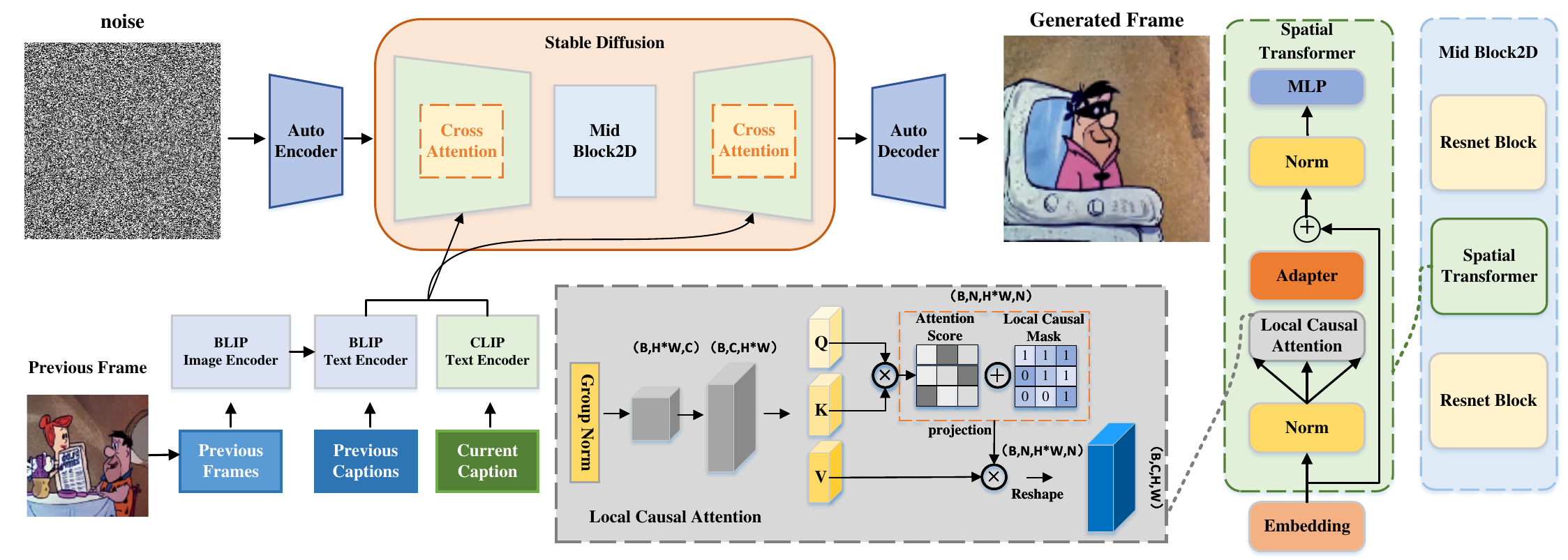}
\caption{Model architecture of  Causal-Story. Our model is inspired by \cite{ARLDM}. The solid line box represents the overall structure of the denoising U-Net section of stable diffusion model, while the dashed line box introduces the specific composition of key modules. The green dashed box displays the location of the local causal attention module and adapter, while the gray dashed box displays the details of the local causal attention module.}
\vspace{-0.5cm}
\label{fig:Model Architecture}
\end{figure*}
\vspace{-0.1cm}
\subsection{Diffusion Processes}
\label{ssec:subhead}
The denoising diffusion probability model \cite{ddpm} consists of a forward process and a reverse process. The forward diffusion process converts the original highly structured and semantically related key point distribution into a Gaussian noise distribution. In the reverse process, the diffusion model learns the required data samples from noise through the UNet\cite{unet} structure. The latent space diffusion model\cite{latentdiffusion} utilizes a pre-trained autoencoder (including an encoder $\mathcal{E}$ and a decoder $\mathcal{D}$ ) to perform the forward and reverse processes of the denoising diffusion probability model in the latent space.

The forward diffusion process converts the original highly structured and semantically relevant key points distribution into a Gaussian noise distribution. In particular, $x$ in this paper denotes latent representations instead of pixels. 

The reverse process of diffusion models is learning the desired data samples from the noise through a UNet\cite{unet} structure. In the reverse process,  we sample from a Gaussian noise distribution $ p(x_t)$. In latent space, the text description is encoded into a latent variable $z$, and $\theta$ are the parameters of the denoising process. The reverse diffusion process can be written as follows:
\begin{equation}
p_{\boldsymbol{\theta}}\left(\mathbf{x}_{t-1} \mid \mathbf{x}_t, \mathbf{z}\right)=\mathcal{N}\left(\mathbf{x}_{t-1} \mid \boldsymbol{\mu}_{\boldsymbol{\theta}}\left(\mathbf{x}_t, \mathbf{t}, \mathbf{z}\right), \beta_t \boldsymbol{I}\right)
\end{equation}
where $p_{\boldsymbol{\theta}}\left(\mathbf{x}_{t-1} \mid \mathbf{x}_t, \mathbf{z}\right)$ represents the reverse transitional probability of key points from one step to the previous step, $\mu_{\boldsymbol{\theta}}\left(\mathrm{x}_t, \mathrm{t}, \mathrm{z}\right)$ is the target we want to estimate by a neural network. $t$ is the timestep indicating where the denoising process has been conducted, which is encoded as a vector based on the cosine schedule\cite{ddpm_improved}. 

\subsection{Local Causal Attention Mask (LCAM)}
\label{ssec:subhead}
To generate consecutive frames similar to stories, we not only need to consider the characteristics of the current caption but also the images generated in the previous frame and their corresponding captions. 
The key to designing a powerful story synthesis model is to enable it to understand the causal relationship between historical captions, frames, and the current caption. We propose a local causal attention mechanism in the model, which enables the model to combine previous captions and frames better to generate the current frame while eliminating confusion effects through a local causal attention mask(LCAM).

The AR-LDM\cite{ARLDM} utilizes an N-to-N self-attention module to uniformly flatten all historical captions and frames into conditional memory using a chain rule, thereby improving the coherence of story generation. However, in the process of story visualization, not all previous frames and captions are related to the generation of the current frame, and longer historical captions often interfere with each other, ultimately reducing the quality of the current frame generation.
According to this conjecture, longer token captions tend to disturb each other because of the confused attention across frames. To alleviate this problem, we propose introducing a local causal masking mechanism so that the learned causal attention module can better adapt to the cases involved in coherent story synthesis with long and complex captions.

We define $L$ as the length of certain story, let $C=[c_1,c_2...,c_L]$ represent the captions of frames, and $F=[f_1,f_2,...,f_L]$ indicate the frames to be generated.  Each caption $c_t$ is corresponding to a frame $f_t \in \mathbb{R}^{C \times H \times W}$, which $t \in (1,L)$. The encoded features that combine both text and image modalities from previous captions and generated frames can be defined as $m_{<t}$
\begin{equation}m_{<t}=\sum_{n=1}^{t-1} B L I P\left(c_n, f_n\right)\end{equation}

where  BLIP\cite{blip} is pre-trained using vision-language understanding and generation tasks with large-scale,  filtered, and clean web data.
We adopt the causal attention mask strategy to achieve this, the attention $CA_t$ of an input feature $m_t$ is calculated via
\begin{equation}
    \resizebox{.9\hsize}{!}{$
    \mathbf{CA}_t=\operatorname{Attention}\left(\mathbf{Q}_t, \mathbf{K}_t, \mathbf{V}_t\right)=\operatorname{softmax}\left(\frac{\mathbf{Q}_t\mathbf{K}_t^{\top}}{\sqrt{d}}+\mathbf{M}\right) \mathbf{V}_t$}
\end{equation}
where $\mathbf{Q}_t, \mathbf{K}_t, \mathbf{V}_t$ are linearly projected features from $m_{<t}$, $d$ denotes the head dimension, and $M$ is a lower triangular matrix ( if $i>j$, $\mathbf{M}_{i, j}=0$ else $\mathbf{M}_{i, j}=-\infty$) during training. 

For coherent story synthesis during inference, the mask is modified to ensure the present token is only affected by the previous tokens with  size $L_M$. We can consider $L_M$ as the size of maximum temporal receptive field. With the help of the causal attention mask, the self-attention layers can be aware of different lengths of tokens, making the causal receptive field adjustable. It can thus effectively mitigate the quality degradation and temporal inconsistency problem for coherent story synthesis.

Moreover, the proposal of the causal attention mask not only improves the cross-frame coherence and the quality of image generation in the continuation and visualization tasks of story visualization but also allows the model to ignore previous parts unrelated to the current frame generation, thereby improving training speed. Specifically, we compared it with AR-LDM in Experiments 3.2.

\subsection{Parameter-Efficient Tuning Utilizing Adapter}
Training Causal-Story from scratch can often be expensive in terms of time and computational resources. To overcome this, we propose an adapter, which is a lightweight module that can fine-tune a pre-trained model with less data. Rather than learning new generative abilities, the module learns the mapping from control information to internal knowledge in Causal-Story. This approach can help achieve efficient parameter tuning without the need for full training.

\subsection{Training Processes}
\label{ssec:subhead}
In the training process, we maximize the log-likelihood\cite{ae} of the model prediction distribution under the actual data distribution to obtain  $\mu_{\boldsymbol{\theta}}$. The training loss can be expressed as the cross entropy of  $p_\theta(x_0)$ optimized under $x_0 \sim q\left(x_0\right)$.
\begin{equation}\mathcal{L}=\mathbb{E}_{q\left(x_0\right)}\left[-\log p_\theta\left(x_0\right)\right]\end{equation}
We can use variational lower bound to approximate the intractable marginal likelihood 
\begin{equation}
\resizebox{.9\hsize}{!}{$\mathrm{E}_{q\left(\mathbf{x}_0\right)}\left[-\log p_\theta\left(\mathbf{x}_0\right)\right] \leq \mathrm{E}_{q\left(\mathbf{x}_{0: T}\right)}\left[-\log \frac{p_\theta\left(\mathbf{x}_{0: T}, \mathbf{z}\right)}{q\left(\mathbf{x}_{1: T}, \mathbf{z} \mid \mathbf{x}_0\right)}\right]$}
\end{equation}
The objective of this process is similar to DDPM\cite{ddpm} except for including text embedding $z$. 
The simplified training loss can be written as a denoising objective:
\begin{equation}\mathcal{L}=\mathbb{E}_{\mathbf{x}_0, \boldsymbol{\epsilon} \sim \mathcal{N}(0,1), t}\left[\left\|\epsilon-\epsilon_\theta\left(\mathbf{x}_t, t\right)\right\|^2\right]\end{equation}
where $\epsilon$ is the noise sampled from standard Gaussian distribution, $\epsilon_\theta\left(\mathrm{x}_{\mathbf{t}}, t\right)$ is the output of the noise prediction model.

During inference, \cite{class} presents classifier-free guidance to obtain more relevant generation results while decreasing sample diversity in diffusion models:
\begin{equation}\hat{\boldsymbol{\epsilon}}=w \cdot \epsilon_\theta\left(\mathbf{x}_t, \varphi, t\right)-(w-1) \cdot \epsilon_\theta\left(\mathbf{x}_t, t\right)\end{equation}
where $w$ is the guidance scale, $ \varphi$ denotes the condition.

\begin{figure}[t]
\centering
\includegraphics[width=3.2in]{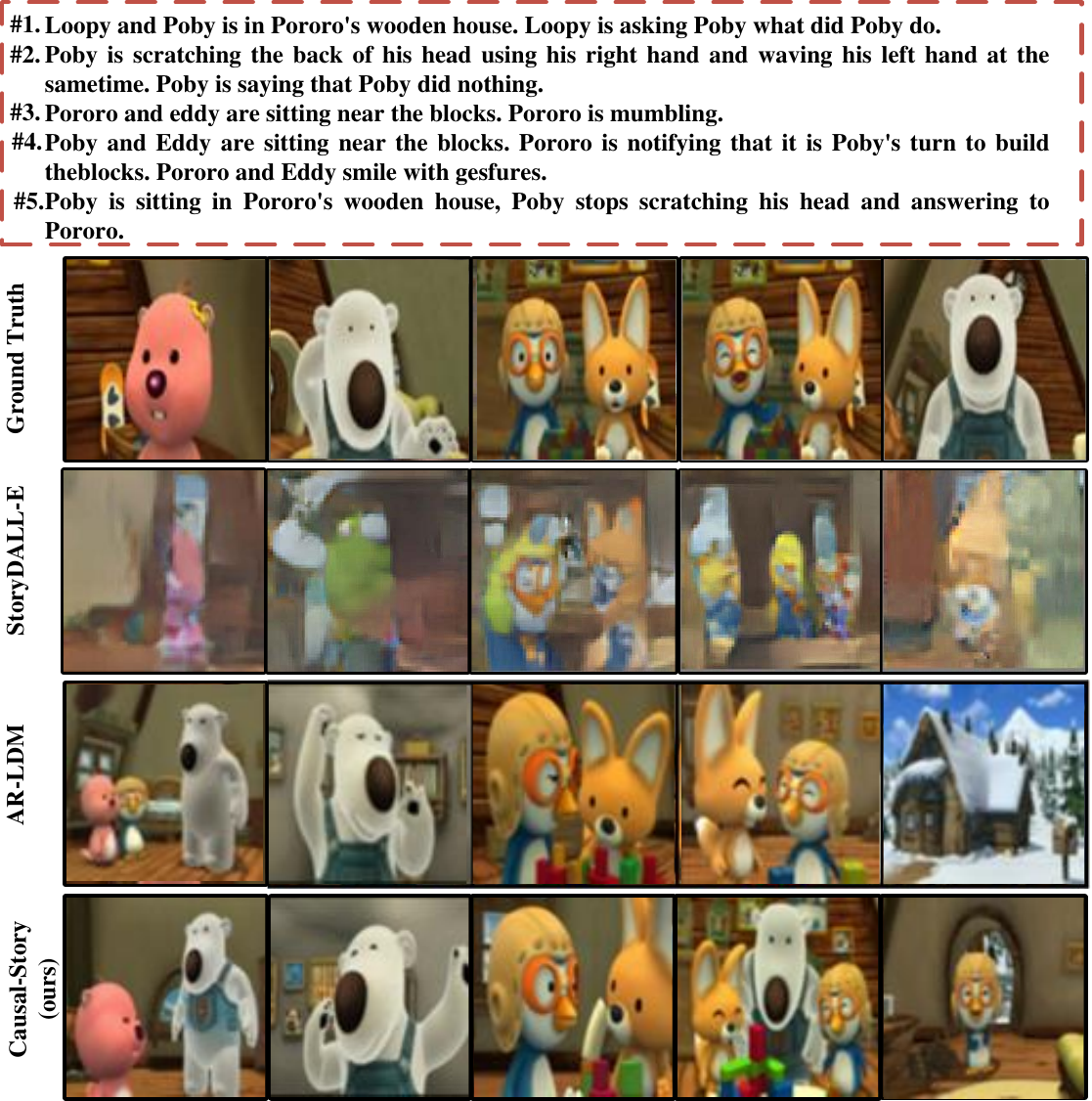}
\vspace{-0.2cm}
\caption{Example of generated images from previous model StoryDALL-E, AR-LDM and our model}
\label{introduction}
\end{figure}

\section{EXPERIMENTS}
\label{sec:majhead}

In section 3.1, we introduced the dataset and our experimental setup. Subsequently, we conducted a comparison of our model and the state-of-the-art technique, considering multiple aspects. In section 3.2, we performed ablation experiments to evaluate the individual advantages of each proposed architecture component.

\subsection{Comparison with the State of the Arts}
\label{ssec:subhead}
Our study involved conducting experiments on two tasks: story visualization and story continuation. Story visualization involves generating a sequence of images that corresponds to a sequence of captions forming a narrative. On the other hand, story continuation is a variant of story visualization that involves using an initial ground truth image as input.PororoSV\cite{storygan} and FlintstonesSV\cite{flitstones} Dataset is used in our experiments. 
\begin{table}\caption{Results on the test sets of PororoSV and FlintstonesSV datasets from various models.\\}
\vspace{-0.3cm} 
\centering
\resizebox{7.5cm}{2.0cm}{
\begin{tabular}{c|c|c|c}
\toprule
\hline
Task   &\multicolumn{1}{c|}{Story Visualization}  &\multicolumn{2}{c}{Story Continuation} 
\\
\toprule  \hline\text{Model}  & \text {PororoSv}  & \text {PororoSv} & \text{FlintstonesSV}\\
\hline \text {StoryGAN\cite{storygan}}          &158.06  &- & - \\
\hline \text {CP-CSV\cite{cpcsv}}          &149.29  & -&-  \\
\hline \text {DUCO-StoryGAN\cite{duco}}          &96.51 & -&-  \\
\hline \text {VLC-StoryGAN\cite{vlcstorygan}}          &  84.96& -&- \\
\hline \text {StoryGANc\cite{storydalle}}          &-  &74.63 & 90.29\\
\hline {VP-CSV\cite{vpcsv}}  &56.08 &-&- \\
\hline\text {StoryDALL·E \cite{storydalle}}      &65.61		&25.9&26.49\\
\hline\text { AR-LDM \cite{ARLDM}}       &16.89 &17.40 &19.38  \\
\hline\text { Causal-Story(ours)}        & \textbf{16.28}		&\textbf{16.98}&\textbf{19.03}\\
\hline
\end{tabular}}
\label{compare-table-itsd}
\end{table}


We evaluated the performance of Causal-Story in terms of story visualization and continuation. FID score is a measure of the distance between the distributions of real and generated images. A lower FID score indicates higher synthesis quality. As depicted in Table 1, Causal-Story achieved a series of new state-of-the-art FID scores on PororoSV and FlintstonesSV datasets.

In addition, we show an example on the PororoSV dataset in Fig. 3.  We can observe that our model is able to maintain text-image alignment and consistency across images. Compared to StoryDALL-E, our model and AR-LDM have significantly improved the quality of generated images. Compared to AR-LDM, our model can better understand text's semantic information and logical relationships. For the first frame generation, AR-LDM mistakenly understood the ``Pororo's house'' in the caption. In the generation of the fourth frame, AR-LDM ignored the character ``Poby'' mentioned in the caption. For the generation of the fifth frame, AR-LDM focused on the exterior image of the ``Pororo's wooden house'' while ignoring the core semantics of captions.

\subsection{Ablation Studies}
\label{ssec:subhead}
In order to analyze the proposed local causal attention and the adapter mechanism, we conducted two ablation studies in this section. Table 1 shows that our model with local causal attention can achieve better FID scores compared to AR-LDM. Meanwhile, according to Fig.3, Causal-Story can better learn causal relationships between contexts and is not affected by irrelevant captions. Furthermore, Table 2 showcases the improvement in model training and sampling speed with the inclusion of the adapter.
\begin{table}[h]
\vspace{-0.2cm} 
\caption{Comparison of Training and Sampling Speeds}
    \renewcommand{\arraystretch}{0.8}\centering
    \centering
    \begin{tabular}{llcc}
        \toprule
         & Train(50 epochs)  & Sample \\
        \midrule
        AR-LDM  & 71h 43m 54s   & 59h 04m 32s   \\
        Causal-Story  & \textbf{65h 31m 38s}   & \textbf{58h 27m 21s}   \\
        \bottomrule
        \vspace{-0.2cm} 
    \end{tabular}
\end{table}

\section{CONCLUSION}
\label{sec:prior}
\vspace{-0.2cm}
Our work applies latent diffusion models to generate coherent images based on textual descriptions. We have designed a local causal attention module that allows the model to learn the causal logical connections between the previous and current frames and captions. We evaluated FID Score on the PororoSV and FlintstonesSV datasets. Researching the visualization results, the coherent story visualization generated by Causal-Story performs well in terms of coherence and image quality. We also found that our method can perform faster training and sampling compared to previous methods with the same number of parameters.

\vfill\pagebreak
\bibliography{refs}
\end{document}